\documentclass[10pt,twocolumn,letterpaper]{article}

\usepackage{cvpr}
\usepackage{times}
\usepackage{epsfig}
\usepackage{graphicx}
\usepackage{amsmath}
\usepackage{amssymb}

\usepackage[breaklinks=true,bookmarks=false]{hyperref}

\cvprfinalcopy 

\def\cvprPaperID{****} 

\begin{document}
	\pagenumbering{gobble}

	
	\title{Attention-based Natural Language Person Retrieval}

	\author{Tao Zhou$^1$, Muhao Chen$^1$, Jie Yu$^2$, Demetri Terzopoulos$^1$\\
		University of California, Los Angeles$^1$\\
		SAIC Innovation Center, San Jose$^2$\\
		{\tt\small \{taozhou, muhaochen, dt\}@cs.ucla.edu}; 
		{\tt\small jerry.j.yu@gmail.com}
	}

	\maketitle
	
	\begin{abstract}
		Following the recent progress in image classification and captioning using deep learning, we develop a novel natural language person retrieval system based on an attention mechanism. More specifically, given the description of a person, the goal is to localize the person in an image. To this end, we first construct a benchmark dataset for natural language person retrieval. To do so, we generate bounding boxes for persons in a public image dataset from the segmentation masks, which are then annotated with descriptions and attributes using the Amazon Mechanical Turk. We then adopt a region proposal network in Faster R-CNN as a candidate region generator. The cropped images based on the region proposals as well as the whole images with attention weights are fed into Convolutional Neural Networks for visual feature extraction, while the natural language expression and attributes are input to Bidirectional Long Short-Term Memory (BLSTM) models for text feature extraction. The visual and text features are integrated to score region proposals, and the one with the highest score is retrieved as the output of our system. The experimental results show significant improvement over the state-of-the-art method for generic object retrieval and this line of research promises to benefit search in surveillance video footage.
	\end{abstract}
	\section{Introduction}
	
	Video surveillance cameras have been deployed in many places---in stores and homes for indoor monitoring, as well as in roadways and parking lots for wide-area observation. Due to the ubiquity of these devices, enabling more effective technologies for surveillance data analysis has become an urgent challenge. This demands automatic recognition and characterization of persons in the large quantities of images/videos, and also requires machines to make such visual information of persons compatible with human understanding. Hence, it is highly desirable for a system to match visual objects with corresponding human language descriptions. The understanding of visual contents is extremely challenging due to factors such as low resolution, deformation, and occlusion, and so is the understanding of human language expressions where semantics are latent and hard to quantify.\par
	
	Several recent papers report promising progress in this area.
	Feris et al.~\cite{feris2014attribute} retrieved human faces from video streams given classified and weighted attributes, but they could not handle contexts like people and scenes, as well as language inputs. Moreover, Socher et al.~\cite{DBLP:journals/corr/abs-1301-3666} evaluated if the image contains seen or unseen classes, whereas Lazaridou et al.~\cite{lazaridou2014wampimuk} focused on predicting keywords for images. The common problem of these efforts is that they rely on traditional language models that are not able to handle sentences with rich semantics and contexts. With the resurrection of neural networks, deep learning has now become popular in visual recognition and natural language processing tasks. Nevertheless, an attentive natural-language-driven retrieval that localizes the most worthwhile objects, including persons, still remains largely unsupported.
	
	In this paper, we address the following problem: Given an image of multiple persons and a sentence that describes a person, we want to localize the most relevant person in the image based on both the text description and the visual features. For example, to retrieve the rider (red rectangle in Figure~\ref{fig:bb}D) given a description saying ``an elderly man wearing a light hat, plaid shirt and dark pants riding a bike'' and attributes such as ``male'' and ``on the right side'', one must look at all the three persons to decide which one best matches the description. In fact, a similar work has been implemented in \cite{6818956} by jointly processing images and text mainly relying on pre-trained objects and knowledge graphs. In contrast to previous works, we develop a deep learning framework to learn a scoring function that takes the text query, attributes and images as input, and outputs scores for the candidate regions. In practice, this line of research motivates many applications, such as the retrieval of suspects and missing persons, when the images or videos of the target person are hidden in the surveillance data. In this case, the police can rely only on the witness's text description.
	
	\section{Related Work}
	
	We discuss the following three lines of related research.
	
	\textbf{Image Captioning and Visual Question Answering.} Image captioning models take an image as input and output a caption describing the contents of the image. These models mainly combine Convolutional Neural Networks for visual feature extraction and Recurrent Neural Networks for caption generation \cite{DBLP:journals/corr/DonahueHGRVSD14,7534740,DBLP:journals/corr/KirosSZ14}. However, it is uncertain that, to what extent these captioning models understand the image. Xu et al.~\cite{DBLP:journals/corr/XuBKCCSZB15} proposed an attention model and showed that the captioning model can zoom into specific regions of the image.
	
	The performance of visual question answering tasks is considered a proxy of the capability of deep neural models for jointly reasoning across the linguistic and visual inputs. 
	Sophisticated approaches have been developed \cite{DBLP:journals/corr/FukuiPYRDR16,DBLP:journals/corr/LuYBP16}, which substantially improve the results in
	performance of visual question answering on the VQA datasets~\cite{DBLP:journals/corr/FukuiPYRDR16}. Nevertheless, Goyal et al.~\cite{DBLP:journals/corr/GoyalKSBP16} demonstrated that simply memorizing 
	question-answer pairs leads to good performance on many visual QA 
	challenges, while simple features, like bag-of-words-based text representations, can perform competitively against many sophisticated approaches~\cite{Jabri2016}.
	
	\textbf{Natural Language Object and Image Retrieval.} Natural language object retrieval aims to localize a target object within an image based on a natural language query for the object. Given a set of candidate object regions, Guadarrama et al.~\cite{Guadarrama14:OOR} generated text from those candidates represented as bag-of-words and compared the word bags to the query text. Other methods generated visual features from query texts and matched them with image regions; e.g., through a text-based image search engine \cite{Arandjelovic12b}. Our work belongs to this category; however, our approach focuses only on the people in urban street scenes. 
	
	Similarly, text-based image retrieval systems select from a set of images the one that best matches the query text. The best match is chosen from a ranking function, which is learned via a Recurrent Neural Network \cite{DBLP:journals/corr/DonahueHGRVSD14,DBLP:journals/corr/MaoXYWY14a}.
	
	\textbf{Deep Attention Models.} The deep attention mechanism was first introduced for machine translation tasks \cite{DBLP:journals/corr/BahdanauCB14}. An extra softmax layer was added to generate weights for the individual words of the sentence, and the quality of the attention or alignment was visualized. 
	
	Due to its enormous impact, the attention mechanism was adopted in other domains. For image captioning, Xu et al.~\cite{DBLP:journals/corr/XuBKCCSZB15} attended on 2D feature maps generated by a Convolutional Neural Network. Similarly for the object retrieval task, Rohrbach et al.~\cite{DBLP:journals/corr/RohrbachRHDS15} learned attention on the relevant image regions in order to reconstruct the input phrase. As for visual question answering, \cite{DBLP:journals/corr/XuS15a,DBLP:journals/corr/ZhuGBF15} proposed models that also attend to image regions or questions when generating an answer.
	
	It is worth mentioning that Lu et al.~\cite{DBLP:journals/corr/LuYBP16} proposed a novel mechanism that jointly learns attention on visual objects and text expression of questions for VQA tasks; that is, the image representation is used to guide the question attention and the question representations are used to guide image attention. However, Das et al.~\cite{DBLP:journals/corr/DasAZPB16} analyzed the consistency between human and deep network attention in visual question answering. The experiment showed that previous attention models in VQA do not seem to target the same regions as humans. As for humans, we usually concentrate on several small regions rather than distribute the attention over the entire image as reflected by deep network attention. Inspired by this discovery, we constructed a visual attention map based the output of Faster R-CNN~\cite{DBLP:journals/corr/RenHG015}.
	
	\section{Data Collection}
	
	The inputs to the neural network are composed of an image, region proposals from the image, a text query, and attributes of target people. Given a target person in the image, the annotation of corresponding text queries and attributes is crowdsourced to Amazon Mechanical Turk (AMT).\footnote{https://www.mturk.com/} Region proposals for 'people' are then generated by a region proposal network in Faster R-CNN. In this section, we first present how the 'person' images were collected.
	
	\begin{figure} [h]
		\centering
		\includegraphics[width=8cm,height=130pt]{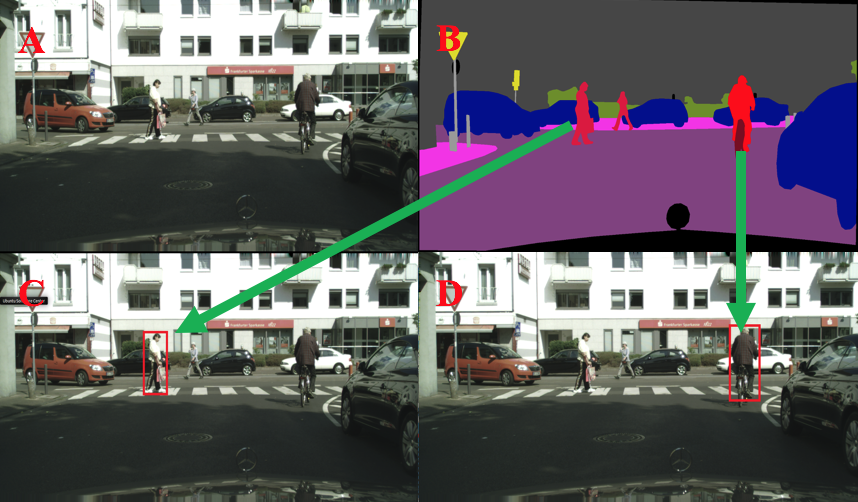}
		\caption{An example of ground truth bounding box generation. A) An original image provided by the CITYSCAPES dataset; B) A corresponding image with segmentation masks provided by the CITYSCAPES dataset; C) a 'Person' with bounding box generated from the segmentation mask; D) a 'Rider' with bounding box generated from the segmentation mask.}
		\label{fig:bb}
	\end{figure}
	
	\subsection{CITYSCAPES Dataset Annotation}
	
	The first challenge of our project was the lack of a dataset for the task of natural language person retrieval. Thus, we turned to the CITYSCAPES dataset \cite{DBLP:journals/corr/CordtsORREBFRS16}, a large-scale benchmark dataset for pixel-level and instance-level semantic labeling. Since the focus of our project is on person retrieval rather than semantic segmentation, only segmentation masks belonging to 'person' and 'rider' categories are transformed into ground truth bounding boxes based on the masks' maximum and minimum value of $(x, y)$ coordinates. Specifically, $(x_{MAX}, y_{MAX})$ denotes the bottom-right corner of the bounding box, while $(x_{MIN}, y_{MIN})$ denotes the top-left corner (Figure~\ref{fig:bb}).
	
	Next, we obtain descriptions and attribute annotations on highlighted people from the images via crowdsourcing. Here we select only bounding boxes with sizes over 5,000 pixels, because small people with low resolution in the image hinder AMT workers from making fine-grain annotation. After the size thresholding step, we keep refining the dataset by removing 1) persons that appear partially on the edge of images; 2) multiple persons within one bounding box in a crowd; 3) blurred persons reported by AMT.
	
	\begin{figure} [h]
		\centering
		\includegraphics[width=8cm,height=150pt]{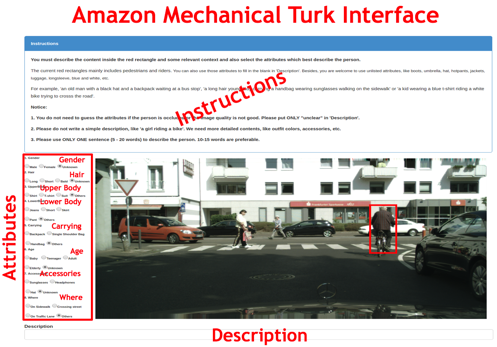}
		\caption{Amazon mechanical Turk (AMT) interface. The AMT workers are given instructions and an image with a target 'people' contained in a red rectangle. They are required to fill in the description and select attributes. The 25 attributes are categorized into 8 classes including: 1) Gender: Male, Female; 2) Hair: Long hair, Short hair, Bald; 3) Upper Body: Shirt, T-shirt, Suit; 4) Lower Body: Jeans, Short, Skirt, Pant; 5) Carrying: Backpack, Single Shoulder Bag, Handbag; 6) Age: Baby, Teenage, Adult, Elderly; 7) Accessories: Sunglasses, Head-phones, Hat; 8) Location: On right side, On left side, In the center.}
		\label{fig:amt}
	\end{figure}
	
	We have designed an interface (Figure~\ref{fig:amt}) for the AMT workers to provide the descriptions and select attributes that best match the appearance of a person inside a bounding box. As for the description, the workers are instructed to use 5--20 words depending on the complexity of the scenario. As for attributes, there are totally 25 values categorized into 8 classes: 'Gender', 'Hair', 'Upper Body', 'Lower Body', 'Carrying', 'Age', 'Accessories', and 'Where'.
	
	The attributes for one person labeled by multiple AMT workers will be settled based on voting during the subsequent data preparation. For example, for the 'Gender' category, if two AMT workers label a person as 'male', while the third one selects 'female', then the final property will be 'male'. For soundness, as long as one worker selects 'unknown' for one category, attributes in that category will be treated as unknown regardless of the other workers' selections.
	
	\subsection{Region Proposal Generation}
	
	\begin{figure} [h]
		\centering
		\includegraphics[width=8cm,height=160pt]{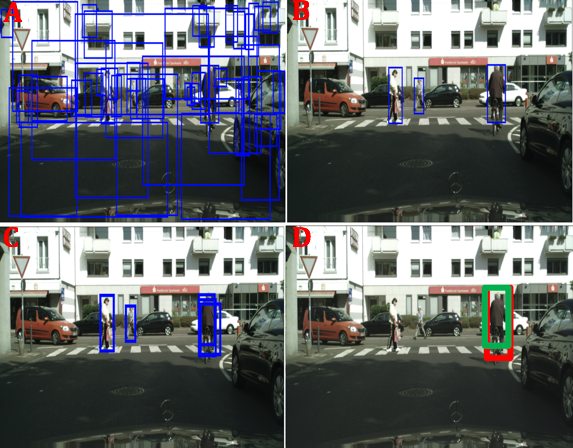}
		\caption{Procedures on region proposals generation for training and testing purpose. A) Blue rectangles are bounding boxes generated by the region proposal network in Faster R-CNN for the category 'people'; B) Region proposals are selected from bounding boxes with confidence beyond 0.5 and size larger than 5000; C) The training dataset is augmented by randomly selecting 3 shifted region proposals whose IOU with ground truth bounding boxes obtained from segmentation masks are beyond 0.5; D) Given a description 'an elderly man wearing a light hat, plaid shirt and dark pants riding a bike', attributes 'male' and 'on the right side', and the region proposals shown in B, the person is retrieved (green region proposal). The ground truth bounding box is shown in red.}
		\label{fig:region-proposals}
	\end{figure}
	
	\begin{figure*}
		\centering
		\includegraphics[width=16 cm,height=190pt]{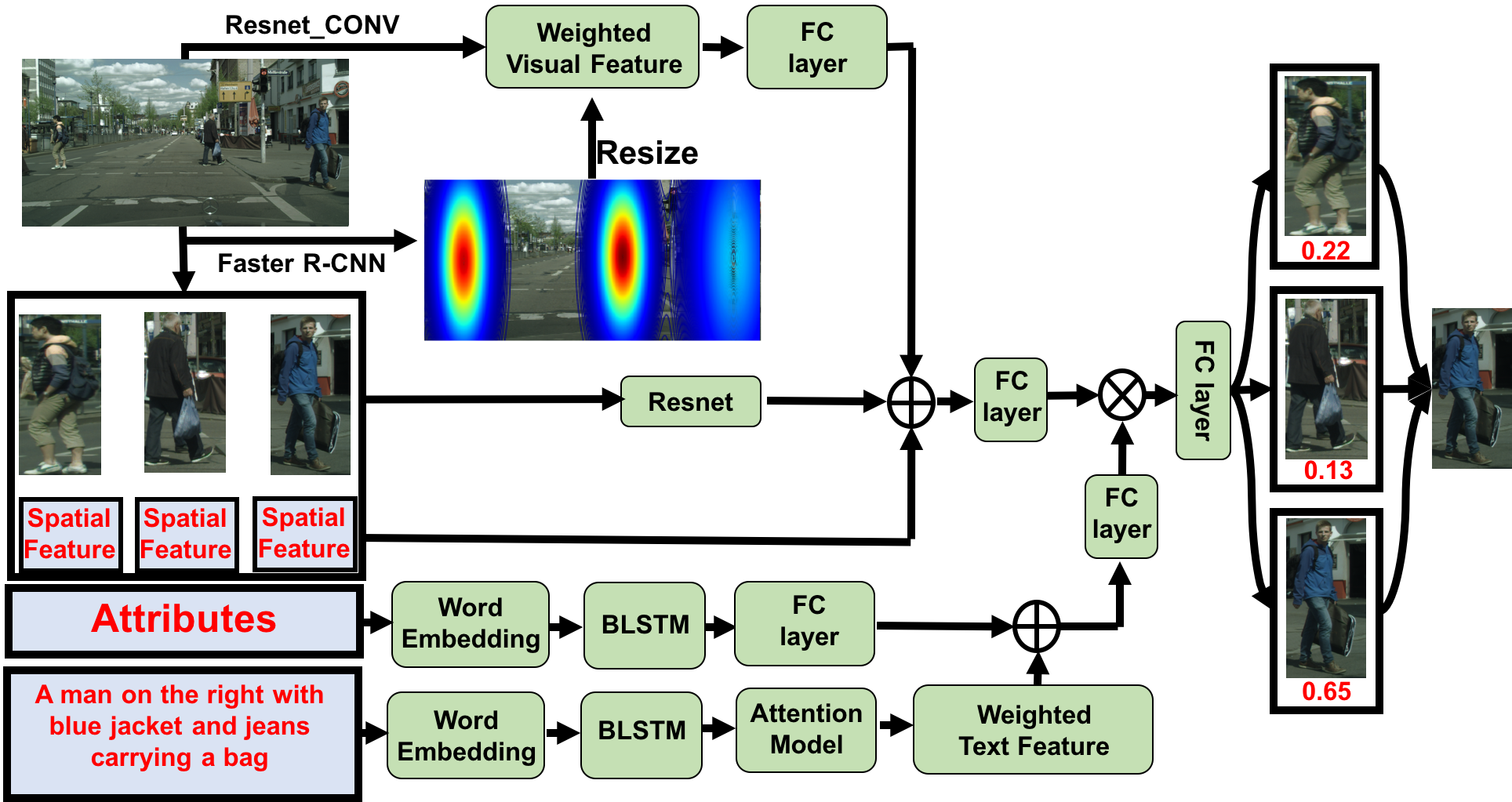}
		\caption{Overview of the attention-based natural language person retrieval framework. The system is given an image, attributes and a text query as input. The region proposals for person are extracted by Faster R-CNN, which also ouputs visual attention probability. The spatial features and the visual features for the whole image and region proposals are concatenated as the final visual features while the attributes and the text query are embedded and then scanned by bidirectional LSTMs to generate text features. The final person retrieval result is output by a fully connected layer whose input is the element-wise multiplication of the concatenated visual and text features. Note: BLSTM means Bidirectional LSTM and FC layer means fully connected layer. All fully connected layers except the last one are used to transform features for concatenation and element-wise multiplication. The operators $\oplus$ and $\otimes$ denote feature concatenation and element-wise multiplication, respectively.}
		\label{fig:framework}
	\end{figure*}

	The region proposal network (RPN) in Faster R-CNN \cite{DBLP:journals/corr/RenHG015} trained on Microsoft COCO \cite{DBLP:journals/corr/LinMBHPRDZ14} is adopted to generate region proposals for people in images. The Microsoft COCO object detection dataset involves 80 categories of objects, such as car and chair, while we are interested only in the detection results on people. A RPN takes an image (of any size) as input, and outputs a set of rectangular object proposals, each with an objectness confidence. The higher the confidence, the more likely it is for the bounding box to contain a person (Figure~\ref{fig:region-proposals}A).
	
	Since most bounding boxes with low confidence do not include the complete imaging of a person, the bounding boxes are filtered by setting the threshold of the confidence to 0.5. As we have mentioned, the minimum size of the bounding box is set to 5,000 pixels in order to avoid small persons in the image (Figure~\ref{fig:region-proposals}B).
	
	To augment the training instances, we expand the dataset by randomly selecting 3 shifted region proposals whose Intersection over Union (IOU) with ground truth bounding boxes is larger than 0.5 (Figure~\ref{fig:region-proposals}C). During the test phase, region proposals without augmentation (Figure~\ref{fig:region-proposals}B) are provided as input to the model for person retrieval (Figure~\ref{fig:region-proposals}D).
	
	Thus far, we have annotated 6,000 'people', 5,000 of which are used for training. A positive training instance is composed of a whole image, a region proposal in the image, its spatial configuration, the corresponding description, and attributes. A whole image, a region proposal along with its spatial configuration and a unrelated description/attributes are paired as a negative sample. Each 'people' has roughly 3 descriptions written by different AMT workers and the positive region proposals were augmented during training by random shift. Here we define the positive-to-negative ratio as 1:1; thus, the total training sample, including positive and negative ones, is around 50,000.
	
	\section{Natural Language Person Retrieval}
	
	Given an image, attributes and a natural language query expression, our goal is to localize the target people. This problem requires both visual and linguistic understanding of the image and the expression. To that end, a model with five main components is proposed: (1) A Convolutional Neural Network to extract local image descriptors and global weighted feature map, (2) word embeddings for text queries and attributes, (3) natural language expression encoder based on bidirectional LSTM networks for both text description and attributes, (4) implicit attention models for images and text queries, (5) a fully connected layer as a classifier. Figure~\ref{fig:framework} shows an overview of our framework.
	
	\subsection{Image Feature Map Extraction}
	
	We apply the L2-norm to the local descriptors at each position on the feature map of region proposals in order to obtain a more robust feature representation. Additionally, the relative coordinates of the region proposal are applied to represent its position in the image. The upper left corner and the lower right corner of the feature map are represented as (-1, -1) and (+1,+1), respectively. The relative center and the relative length of the width and height of the region proposal are also incorporated. Thus, the 8-dimensional spatial features (Figure~\ref{fig:framework}) are [$x_{r-min}$, $y_{r-min}$, $x_{r-max}$, $y_{r-max}$, $x_{r-center}$, $y_{r-center}$, $w_{r-box}$, $h_{r-box}$], where $r$ denotes the relative quantity. In our implementation, this model component takes the refined region proposals generated by Faster R-CNN as the input, and outputs the results by first unifying the size of region proposals to $224 \times 224$, then extracting visual features using Resnet152 \cite{DBLP:journals/corr/HeZRS15} pre-trained on the ILSVRC classification task \cite{imagenet_cvpr09}. The global feature map of the whole image generated by the first convolutional block is multiplied by the attention weights generated from Faster R-CNN to obtain weighted representation of the whole image (see Section~\ref{sec:impl-attent-model} for more details).
	
	\subsection{Word Embeddings}
	\label{sect:we}
	
	As a requisite of natural language processing, the Skip-gram model~\cite{mikolov2013distributed} is used to transform the sentence expressions into matrix representations that are understandable by the neural networks.
	The Skip-gram model represents words in a dense vector space, and closely embeds words with highly similar semantics.
	Typically, the Skip-gram model is trained on a large document corpus $D$, where each word in the vocabulary is first randomly initialized as a $k-$dimensional unit vector.
	Then, Skip-gram maximizes the log-linear energy function defined as
	\begin{equation}
	J=\sum_{(w,c) \in D}(\mathrm{log} \frac{1}{1+e^{-v_c\cdot v_w}}-\sum_{c' \in D} \mathrm{log} \frac{1}{1+e^{v_{c'}\cdot v_w}})
	\end{equation}
	such that $v_w$ is the embedding vector of word $w$, $v_c$ and $v_{c'}$ are the embedding vectors of the contexts $c$ and $c'$.
	By maximizing $J$, the learning process approximates the semantic similarity of words based on their co-occurrences in the local contexts.
	Hence, important semantic features, such as topics, sentiments, objects, and attributes, are often highlighted by the embeddings in the matrix representations of sentences. In our experiment, we pre-trained the Skip-gram model on the entire Wikipedia dump, for which we set the dimensionality $k=300$ and the length of contexts $|c|=20$.
	To enrich the knowledge of the vocabulary, during the preprocessing of the corpus, Wikipedia entities are recognized from the article-based maximum matching, and frequent 2-grams are also considered to mine frequent phrases.
	
	\subsection{Encoding the Descriptive Sentence using an BLSTM Network}
	
	We represent the text description of each image region as a fixed-length sequence of $N$ words. If the text length is larger than $N$, only the first $N$ words are utilized for language feature extraction. Otherwise, the sequence is padded with an empty token $\langle unk \rangle$. Here, each sequence of image description is represented as a matrix using the word embeddings described in Section~\ref{sect:we}. Then we use a bidirectional Long-Short Term Memory (BLSTM) network~\cite{Hochreiter:1997:LSM:1246443.1246450,650093} with a 1,000-dimensional hidden state to scan through the matrix. 
	After the BLSTM network has taken the entire text sequence, the hidden states are concatenated as a single vector $h_t = [h_t^fw\; h_t^bw]$ that encodes the description. The superscript f and b denote forward and backward hidden states, respectively. The bidirectional LSTM gates are computed as
	\begin{align}
	&i_t = S(W_{xi}x_t + W_{hi}h_{t-1} + W_{ci}c_{t-1} + b_i) \\
	&f_t = S(W_{xf}x_t + W_{hf}h_{t-1} + W_{cf}c_{t-1} + b_f)\\
	&c_t = f_tc_{t-1} + i_t\tanh(W_{xc}x_t + W_{hc}h_{t-1} + b_c)\\
	&o_t = S(W_{xo}x_t + W_{ho}h_{t-1} + W_{co}c_{t} + b_o) \\
	&h_t = o_t\tanh(c_t),
	\end{align}
	where $S$ is the logistic sigmoid function, and $i$, $f$, $o$, and $c$ are, respectively, the input gate, forget gate, output gate, and cell activation vectors, all of which are of the same size as the hidden vector $h$.
	
	Additionally, we concatenate all attributes and send it to the bidirectional LSTM to generate the encoding of the attributes. Unlike for the text queries, the attention model is not applied to attributes, since they are equally important. Here we select the bidirectional LSTM rather than the LSTM as a text feature encoder, because the concatenated attributes and text queries have independent sequences. For example, an AMT worker may use two parts of a sentence to describe the upper body first and then the lower body of a person, 
	and the second part might not depend on the first part. The bidirectional LSTM
	scans the text sequence twice in inverse directions (i.e., front to back, then back to front).
	
	\subsection{Implicit Attention Model for Text and Visual Features}
	\label{sec:impl-attent-model}
	
	For the text features, the concatenated state $h_t$ contains information from a word $w_t (t = 1,...,T)$ as well as the context before and after the word $w_t$. Then the attention weights $\alpha_t$ for each word $w_t$ are obtained by a linear projection over $h_t$ followed by a softmax defined as
	\begin{align}
	&\alpha_t = \dfrac{\exp(\beta^Th_t)}{\sum_{t=1}^{T}\exp(\beta^Th_t)}.
	\end{align}
	for which the weighted word expression $h_t^{'}$ is defined as
	\begin{align}
	&h_t^{'} = \alpha_t h_t.
	\end{align}
	
	As for the visual features, the corresponding attention map is generated from Faster R-CNN (the center position, height, and width of region proposals). Here we use a bivariate normal distribution to represent the probability distribution for the $i_{th}$ region proposal:
	\begin{align}
	&P_i(x,y) = \dfrac{1}{2\pi\sigma_{ix}\sigma_{iy}}\exp(-\dfrac{z}{2}),
	\end{align}
	where
	\begin{align}
	&z = \dfrac{(x - \mu_{ix})^2}{\sigma_{ix}^2} +  \dfrac{(y - \mu_{iy})^2}{\sigma_{iy}^2},
	\end{align}
	for which $\mu_{ix}$ and $\mu_{iy}$ are the center position of the $i_{th}$ region proposal, while $\sigma_{ix}$ and $\sigma_{iy}$ are the half width and the half height of the region proposal, $i = 1,...,N$, with $N$ being the number of region proposals of the image.
	
	The final attention map for the whole image is defined as
	\begin{align}
	&P'(x,y) = {1\over N}\sum_{i=1}^{N}P_i(x,y).
	\end{align}
	
	\subsection{Fully-Connected-Layer Classifier}
	
	The combination of the global and local visual features is multiplied to the 
	concatenation of the text and attribute features in an element-wise fashion, 
	which are then fed into a fully-connected-layer classifier.
	
	During the training process, each training instance is a tuple $(I, RP, X, E, A, T)$, where $I$ is a whole image, $RP$ denotes a region proposal in the image, $X$ is the spatial configuration of the region proposal, $E$ is a natural language expression describing the region, $A$ denotes the corresponding attributes for the person inside the region proposal, and $T$ is the tag that marks whether the person inside the region proposal matches the natural language expression and attributes.
	We use the sigmoid cross-entropy loss function:
	\begin{align}
	&\textrm{loss} = -L \log(S(s)) - (1-L) \log(1-S(s)),
	\end{align}
	for which $L$ is the ground truth label (true or false), $s$ is the score output of the neural network, and $S$ is the sigmoid function.
	
	In the test phase, the difference is that all region proposals in the image are sent to the model and the classifier will output a score for each region proposal. The region proposal with highest score is retrieved as the final output of our system (Figure~\ref{fig:framework}).
	
	\section{Performance Evaluation and Analysis}
	
	We performed our experiment using one Nvidia GeForce GTX 1080 Graphics Card. For the visual features, we only train the last fully connected layer of Resnet152 to obtain the local image feature while we generated the global image features separately without fine-tuning any layer of Resnet152.
	
	The results are reported in Table~\ref{tbl:rst}, where the metric 'Rec@1' is the recall of the highest scoring box (the percentage of the highest scoring box being correct), and 'Rec@2' is the percentage of at least one of the 2 highest scoring proposals being correct. Overall, our attention-based natural language person retrieval framework leads to a roughly 35\% increment on Rec@1 as compared to random selection. In fact, the model \cite{hu2016natural} pre-trained on the ReferIt dataset and tested on CITYSCAPES (row 3) is even worse than random selection (row 2), while the model pre-trained on CITYSCAPES increased the accuracy by 5\% (row 4). 
	
	\begin{table} [h]
		\centering
		\begin{tabular}{| l | l | l | l | }
			\hline
			& Trained on & Rec@1 & Rec@2\\ \hline
			Random & &38\% &60\%\\ \hline
			UCB~\cite{hu2016natural} & Referit & 37.3\% &59.7\%\\ \hline
			UCB~\cite{hu2016natural} & CITYSCAPES & 43.7\% & 65.3\%\\ \hline
			Ours & CITYSCAPES & 74.6\% & 85.7\%\\ \hline
		\end{tabular}
		\caption{Test Performance on CITYSCAPES dataset}~\label{tbl:rst}
	\end{table}
	\vspace{-1em}
	
	As compared to \cite{hu2016natural}, the most significant improvement comes from how the visual and text features are combined. Traditionally, the visual and text features are simply concatenated as the input to a multi-layer perceptron to generate a score for person retrieval \cite{zhou2017natural}. However, when the features are multiplied in an element-wise fashion, Rec@1 increases by roughly 10\%. Indeed, this phenomenon has been observed in other works \cite{ba2014multiple,yin2015abcnn} conducting cross-module analysis.
	
	Furthermore, the pre-trained word embeddings with named entity annotation further leverages the Rec@1 by 8\%, in comparison to the one-hot vectors applied in \cite{hu2016natural}. The embedded text and attributes lead to 5\% and 3\% increments of Rec@1, respectively. It is worth mentioning that some persons are extremely hard to be distinguished from others due to similar clothes and appearance. In such cases, the relative position of the person in the image (right, left, or center) plays a major role in constraining the search range. This strategy also leverages Rec@1 by around 7\%.
	
	Due to the attention mechanism, the model has a sense of focus on both images and text queries. The combined effect of attention leads to a 5\%
	increment of Rec@1. Other minor changes, such as fine-tuning the
	hyperparameters and the last few pre-trained layers of Resnet152, also yield improvement.
	
\begin{figure} [h]
		\centering
		\includegraphics[width=8cm]{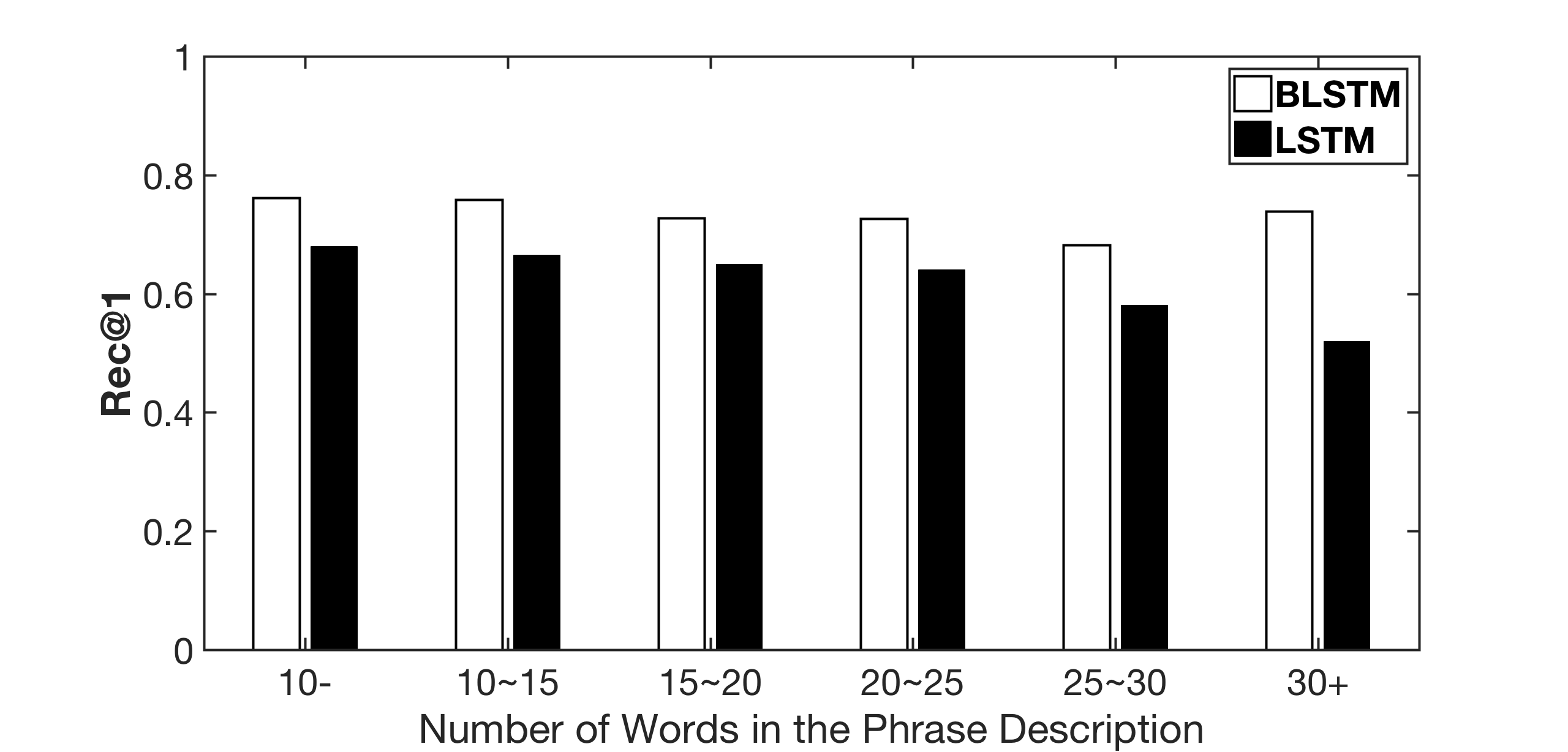}
		\caption{Effect of the Phrase Description Length. The horizontal axis indicates the number of words, while the vertical axis indicates Rec@1.}
		\label{fig:pdl}
	\end{figure}

	We further investigate the contribution of each component to the result. Even though we instructed AMT workers to describe people using 5--20 words, they did not always follow the instructions. Thus, we decided to examine the effect of the phrase description length (Figure~\ref{fig:pdl}). For example, in cases where descriptions are limited to 5--10 words, Rec@1 is 75\% using BLSTM. A description with more than 25 words, however, easily deteriorates the result when LSTM is applied. As a comparison, the accuracy does not show an apparent drop for BLSTM. This demonstrates the effectiveness of bidirectional LSTM on the non-logic dependent description expression.
	
	\vspace{-1em}
	\vspace{-1em}
	
\begin{figure}[h]
		\centering
		\includegraphics[width=8cm]{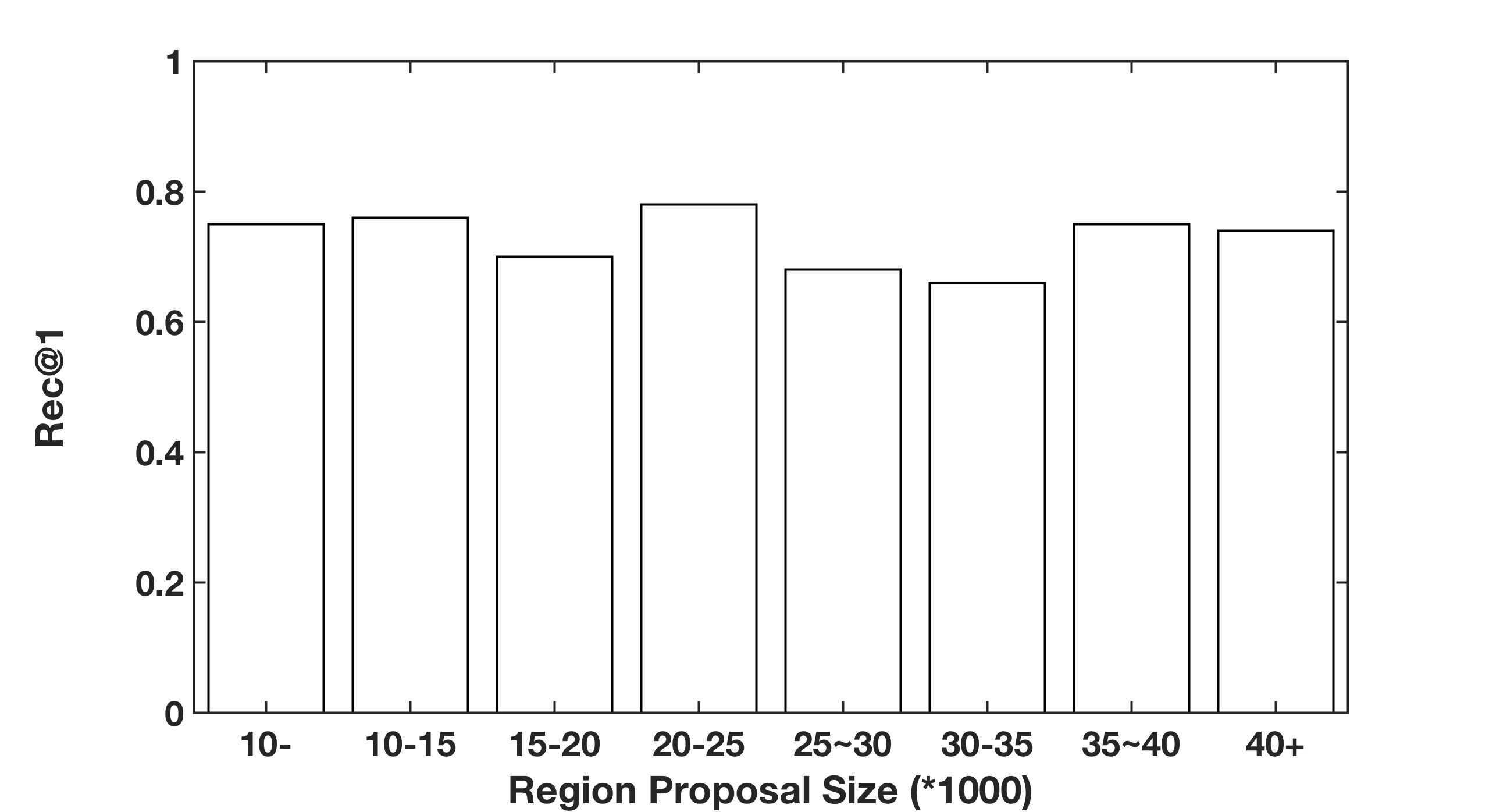}
		\caption{Effect of Region Proposal Size. The horizontal axis indicates the region proposal size while the vertical axis indicates Rec@1.}
		\label{fig:rps}
	\end{figure}

	Figure~\ref{fig:rps} shows the effect of region proposal size. We discovered that the sizes of the region proposals do not show notable differences. Our hypothesis was that all cropped images are resized to $224 \times 224$ as the visual input to get the local visual features. In this way, the network is insensitive to the original size of the cropped image.
	
\begin{figure}
		\centering
		\includegraphics[width=8cm]{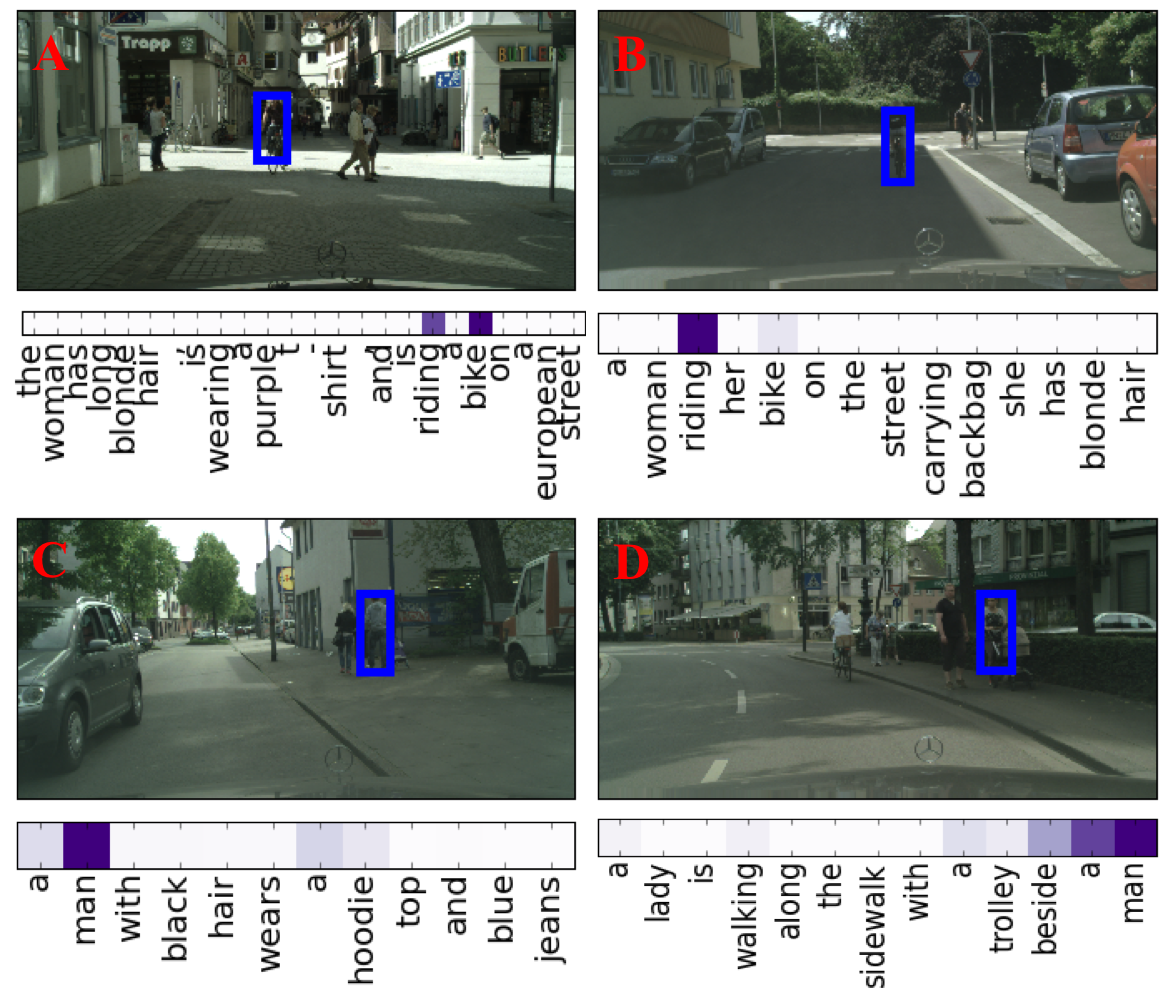}
		\caption{Demonstration of attention mechanism on person retrieval. A) The text query is attended on 'riding a bike'; B) The text query is attended on 'riding her bike'. C) The model focuses on 'man' and 'hoodie'. D) The surrounding information 'beside a man' due to the incorporation of the entire image feature vector.}
		\label{fig:attention}
	\end{figure}

	Finally, the effect of the attention mechanism is examined in Figure~\ref{fig:attention}. Based on the observations, the model has remarkable accuracy on recognizing riders and the attention is usually focused on the words 'riding' or 'bike'. Also, due to the weighted global image features, the attention mechanism allows the model to reason about the surrounding environment.
	
	\section{Conclusion}
	
	In this paper, we presented what was to our knowledge the first attention-based natural language person retrieval system. A large-scale benchmark dataset was constructed using crowdsourcing and processed using Faster R-CNN. A new deep-learning-based framework was further designed to match visual and text representations. Thus, an image, text query, and attributes are the inputs of our framework, which selects a region proposal with the highest score. Compared to the state-of-the-art object retrieval method, a substantial increment in performance was observed in our experiments. In future work, we will investigate a learning-based visual attention model on feature maps from multiple convolution layers which may further improve retrieval performance and compare regions across different images instead of extracting regions from a single image. 
	
	{   \small
		\bibliographystyle{ieee}
		\bibliography{egbib}
	}
	
\end{document}